\documentclass[runningheads]{llncs}
\usepackage{graphicx}

\usepackage{color}
\usepackage{multirow}
\usepackage{flushend}
\usepackage{subfigure}
\usepackage{amsmath}
\usepackage{amssymb}

\begin{document}
\title{Mining Minority-class Examples With Uncertainty Estimates }
\author{Gursimran Singh\inst{1} \and
Lingyang Chu\thanks{Gursimran Singh and Lingyang Chu contribute equally in this work.} \inst{2} \and
Lanjun Wang\inst{3} \and
Jian Pei\inst{4} \and
Qi Tian\inst{5} \and
Yong Zhang\inst{1}}
\authorrunning{Singh and Chu et al.}
\institute{Huawei Technologies Canada, Burnaby, Canada \and
McMaster University, Hamilton, Canada \and
Tianjin University, Tianjin, China \and
Simon Fraser University, Burnaby, Canada \and
Huawei Technologies China, Shenzhen, China
\email{\{gursimran.singh1,yong.zhang3,tian.qi1\}@huawei.com}
\email{chul9@mcmaster.ca}
\email{wang.lanjun@outlook.com}
\email{jpei@cs.sfu.ca}
}
\maketitle

\begin{abstract}
In the real world, the frequency of occurrence of objects is naturally skewed forming long-tail class distributions, which results in poor performance on the statistically rare classes. A promising solution is to mine tail-class examples to balance the training dataset. However, mining tail-class examples is a very challenging task. For instance, most of the otherwise successful uncertainty-based mining approaches struggle due to distortion of class probabilities resulting from skewness in data. In this work, we propose an effective, yet simple, approach to overcome these challenges. Our framework enhances the subdued tail-class activations and, thereafter, uses a one-class data-centric approach to effectively identify tail-class examples. We carry out an exhaustive evaluation of our framework on three datasets spanning over two computer vision tasks. Substantial improvements in the minority-class mining and fine-tuned model’s task performance strongly corroborate the value of our method.

\end{abstract}

\section{Introduction}
\label{sec:intro}
A majority of powerful machine learning algorithms are trained in a supervised fashion \cite{lin2017focal,deng2009imagenet}. Hence, the quality of the model depends heavily on the quality of data. However, in many real-life applications, the distribution of occurrence of examples is heavily skewed, forming a long tail. This has a detrimental effect on training machine learning models and results in sub-optimal performance on underrepresented classes, which is often critical \cite{bengio2015battle,he2009learning}. For instance, in medical diagnosis the inherent data distribution contains a majority of examples from a \textit{common-disease}  (head-class) and only a minority from a \textit{rare-disease} (tail-class, a.k.a. minority-class). Similarly, this problem naturally occurs in many domains like fraud detection, spam detection, autograding systems, and is so pervasive that even carefully crafted research datasets like ImageNet, Google Landmarks, and MS-CELEBS-1M are all class-imbalanced \cite{aggarwal2020active,kazerouni2020active,deng2009imagenet,singh2016question}.

\begin{figure}[t]
\begin{center}
\includegraphics[width=\linewidth]{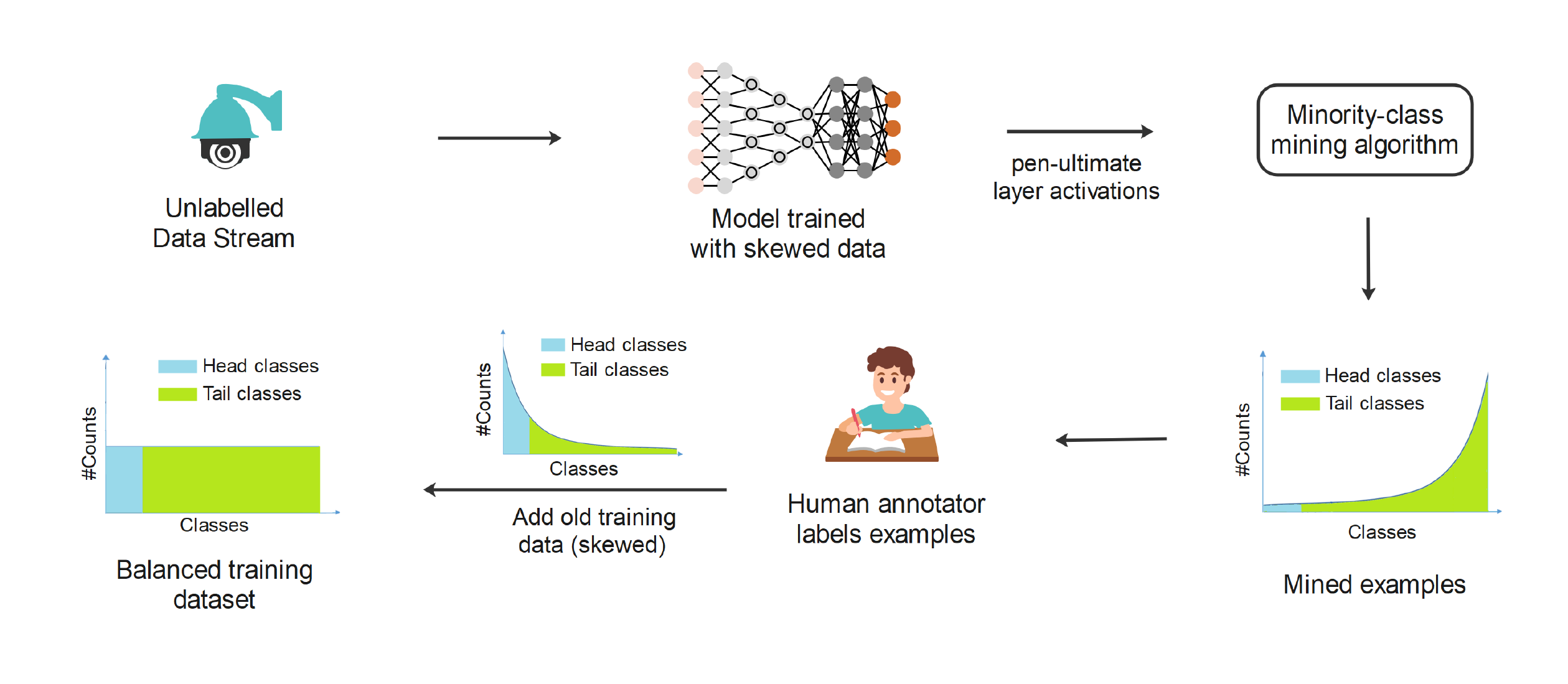}
\end{center}
\caption{An illustration of using minority-class mining approach to balance a skewed training dataset. Minority-class mining algorithm takes penultimate-layer activations corresponding to unlabelled data as input and outputs examples which mostly belong to one of the tail-classes. After the human annotator labels these examples, we combine these with the old training dataset (skewed) to obtain a new balanced training dataset. Thereafter, this new dataset can be used to train effective machine learning models. }
\label{fig:overview}
\end{figure}

In order to balance these datasets, we need to augment them by procuring additional labeled examples from the tail classes.  However, any pool of data sampled uniformly at random is similarly skewed and contains few desired tail-class examples \cite{attenberg2010label}. Consequently, human workers will spend most of their time labeling useless majority-class examples, wasting time and resources. To address these disproportions, we must design intelligent algorithms to actively mine minority-class examples. Such an algorithm has immense practical value in balancing skewed datasets by procuring statistically rare labeled examples in a cost-effective manner. As a result, we can train effective machine learning models in all the above-mentioned cases. \textit{How can we mine minority-class examples from unlabeled datasets automatically?} Given a biased model trained with a pool of skewed data, this work addresses the problem of mining (or active sampling) of minority-class examples from unlabeled data. Refer to Fig. \ref{fig:overview} in for an illustration of this process.

Despite being extremely useful, minority-class mining, unfortunately, is a very challenging task. The difficulty is three-fold. First, although tail-class examples are semantically distinct from head-class examples, it is not easy to separate them in high-dimensional input spaces. Often, examples lie on complex manifolds that are not amenable to unsupervised density-based modeling \cite{thudumu2020comprehensive}. 
Second, even in the output space, the bias in the trained model due to class imbalance subdues tail-class activations, resulting in distortion of the model's uncertainty estimates.
This effect is aggravated in deep neural networks where due to pervasive overfitting of the final softmax layer \cite{guo2017calibration,mukhoti2020calibrating}, head-class activations completely eclipse tail-class activations. Hence, the otherwise successful informative-sample mining methods in active learning literature, which rely on the model's uncertainties in the output space, are well-known to perform poorly in mining tail-class examples \cite{aggarwal2020active,attenberg2010label,attenberg2011inactive,c2018active}.
Third, many of these approaches, like max score \cite{culotta2005reducing}, and entropy \cite{shannon2001mathematical}, aggregate the information-rich class-probability distribution into a scalar, rendering it unable to model relationships among individual values. These metrics are fixed formulae (like $-\sum_{k}^C p^i_k \log (p^i_k)$ for entropy), having no parameters, and unlike a learning process, are unable to adapt to identifying and discriminating patterns in a particular dataset. However, in reality, head-class and tail-class examples exhibit myriad characteristic patterns in the model's output space, which needs to be \textit{learned} using an expressive modeling technique.

In this paper, we address these challenges and propose a simple, yet effective, technique to mine minority-class examples. There are two key ideas of our approach. First, we propose to use pen-ultimate layer activations of the trained model, which, unlike the input or output space, are not afflicted by the high-dimensionality, miscalibration, or overfitting problem. We use these activations to obtain re-calibrated head-class and tail-class probabilities, which retain their distinctive patterns and are more amenable to the learning process.  Second, we formulate the minority-class mining problem as a one-class learning algorithm.
The resultant data-centric method, unlike a fixed formula, helps to \textit{learn} characteristic activation patterns corresponding to majorly present head-class examples and flags aberrant tail-class examples as anomalies. As shown in the experiments section, both of these simple ideas lead to a \textit{substantial improvement} in the minority-class mining performance.

\textbf{Contributions:} Our contributions are three-fold. \textbf{1)} We identified the key challenges and presented a novel framework to effectively mine minority-class examples using the uncertainties in the model's output probabilities. This is in contrast to the previous line of thought that uncertainty-based approaches alone are not effective in mining minority-class examples \cite{aggarwal2020active,attenberg2010label,c2018active}. 
\textbf{2)} Our approach can be easily extended to many machine learning tasks since it only relies on the trained model's output activations. We demonstrate the effectiveness of the proposed method on two different computer vision tasks, namely classification, and object detection. In doing so, we device a methodology to extend tail-class mining to object detection, which, to the best of our knowledge, is the first attempt in this direction. 
\textbf{3)} We devise a new experimental setup to analyze the minority-class mining performance based on the individual class's skewness. We show that our approach has a desirable property of mining a greater number of direly-needed examples from the most skewed classes.

\section{Related Works}
Imbalanced data is a major problem in machine learning \cite{he2009learning,krawczyk2016learning,zhu2014capturing}. Among many possible approaches, a promising direction is to mine tail-class examples with an aim to balance the dataset. In this direction, although uncertainty sampling \cite{culotta2005reducing,shannon2001mathematical,dagan1995committee} is an effective method for mining generic informative examples \cite{culotta2005reducing,settles2009active,shannon2001mathematical}, they are known to be ineffective in mining minority-class examples \cite{aggarwal2020active,attenberg2010label,attenberg2011inactive,c2018active}. This has been attributed to the fact that uncertainty sampling, being biased on previously seen examples, ignores regions that are underrepresented in the initially labeled dataset \cite{kazerouni2020active}. 
Several approaches \cite{chen2011active,kirshners2017entropy,tomanek2009reducing,ertekin2007learning} propose to account for the skewness by using prior information about class imbalance to boost query scores corresponding to tail classes. For instance, \cite{kirshners2017entropy} proposes to weight Shannon's entropy by the inverse of class-proportion ratios in the training dataset. Similarly, \cite{chen2011active} proposes to boost the vote score corresponding to tail classes of a query-by-committee strategy. However, boosting tail-class activations does not help in high-skew environments (especially in overfit neural networks) where tail-class probabilities are very weak or non-existent. 

As a workaround, recent approaches \cite{bhattacharya2019generic,kazerouni2020active} propose augmenting uncertainty sampling with an exploration/ geometry/ redundancy criteria in the input space. The key insight is to allow exploration to new uncertain areas in the input space. They either use bandit algorithms \cite{kazerouni2020active} or linear programming formulation \cite{bhattacharya2019generic} as an optimization objective to choose among different criteria. However, the exploration/ geometry/ redundancy criteria in the input space perform poorly in high-dimensional spaces. Moreover, they incur additional computational or memory overhead (quadratic in the number of examples and classes), making them unscalable to large datasets.
Another set of techniques \cite{attenberg2010label,c2018active} proposes to use generation queries (tasking crowd workers to create examples of tail classes), and claim better performance than mining-based methods. However, these techniques have limited applicability (not possible in the medical domain) and are financially expensive due to increased human effort \cite{c2018active}.

In contrast to the above-mentioned workarounds, we do not rely on any additional input like exploration or example generation by humans. Hence, our method does not suffer from the above-mentioned shortcomings. Instead, our method takes a more heads-on approach and addresses the key challenges of uncertainty sampling (see Section 1) in skewed settings. Further, most of the approaches (except \cite{bhattacharya2019generic}) mentioned above assume a binary classification setting, with a single minority and majority class. Additionally, these are designed with classification in mind and, in many cases, are not easily extendable to other learning tasks. Our approach, on the other hand, directly handles multi-class settings (multiple minority and majority classes) and can be easily extended to other tasks, like object detection.

\section{Our Method}

Consider a machine-learning task where a model is trained using a pool of finite labeled dataset $D_L$. Each instance in $D_L$ is mapped to one or more elements of a set of classes $C$. Additionally, we are given that there is a class imbalance in dataset $D_L$, such that a subset of classes $C_M \subset C$ are relatively under-represented (a.k.a minority classes or tail-classes). 
Given a pool of unlabelled data $D_U$, sampled from a space of instances $\mathcal{X}$, the \textit{minority-class example mining} algorithm aims to learn a ranking function $\Omega_{\beta}(x):\mathcal{X} \rightarrow \mathbb{R}$, such that higher scores indicate a high likelihood of a sample $x$ belonging to one of the minority-classes. In this work, we learn the parameters $\beta$ in a one-class setting using the training dataset $D_L$. Specifically we expect the function $\Omega_{\beta}(x)$ to overfit the majorly-present head-class data in $D_L$ so that unfamiliar minority-class samples can be flagged as anomalies.
Once the tail-class examples are mined, they are sent for labeling. The labeled tail-class examples aid in balancing the dataset and, thereafter, train effective machine learning models in real-life applications with skewed long-tail class distributions.

\begin{figure*}[t]
\centering
\vspace{-4mm}
\includegraphics[width=\textwidth]{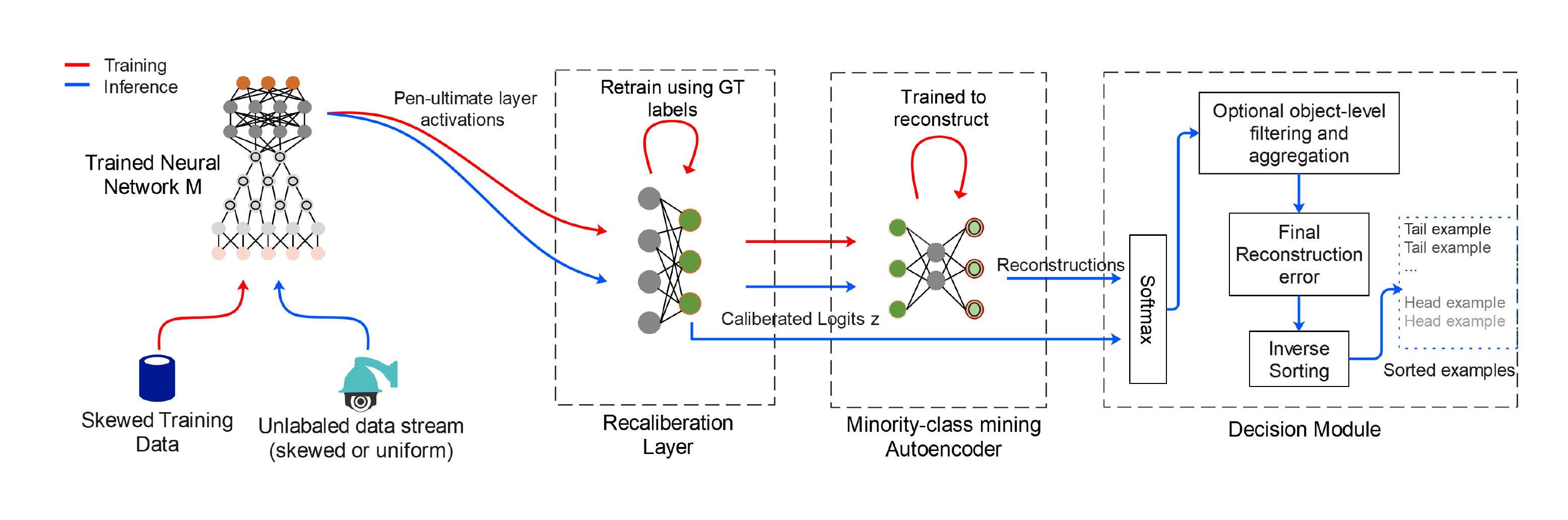}
\vspace{-8mm}
\caption{Overall framework of the proposed method.}
\vspace{-4mm}
\label{fig:paper}
\end{figure*}

The proposed minority-class hard example mining framework (shown in Fig \ref{fig:paper}) consists of three major components: (a) the re-calibration layer (RC layer), (b) the minority-class mining autoencoder (MCM-AU), and (c) the decision module. 
First, we use the re-calibration layer to recalibrate the final layer probabilities which are distorted due to class imbalance and overfitting. Specifically, the recalibration layer takes pen-ultimate layer activations of model $M$ as input and outputs the calibrated logits $z \in R^C$. 
Second,  we model the calibrated logits $z$ using an autoencoder which acts as an anomaly detector to help separate tail-class examples due to their anomalous activation patterns.
Specifically, the calibrated logits $z$, obtained from the recalibration layer, is passed into the autoencoder and a potential reconstruction $\hat{z}$ is obtained as the output. Finally, based on the quality of the reconstruction, instances are flagged as minority-class examples by the decision module. Specifically, it takes the original calibrated logit $z$ and its reconstruction $\hat{z}$ as input, computes the reconstruction error for each example.
The examples with the highest reconstruction error are selected for labeling. We describe these components in more detail as follows.

\subsubsection{The Recalibration Layer:}
The Recalibration layer (RC-Layer) aims to solve the problem of distorted class-probability distribution for tail-class examples. As explained earlier, this problem occurs due to two main reasons: overfitting in the final softmax layer and subdued tail-class activations. The overfitting and the resulting miscalibration in softmax probabilities is a known phenomenon observed in many modern multi-class classification networks \cite{guo2017calibration,mukhoti2020calibrating}. This has been attributed to prolonged training to minimize the negative log-likelihood loss on a network with a large capacity. Subdued activations for tail classes are a result of skewness and insufficient data due to the class imbalance. Lack of sufficient data causes the weak tail-class activations to be dominated by stronger head-class activations.

To mitigate these problems, the overfit softmax layer of the model $f_\theta$ is replaced with a structurally similar newly-initialized layer (recalibration layer). The intuition behind replacing only the last layer is based on the idea that overfitting and miscalibration are mainly attributed to weight magnification particularly in the last layer of the neural network \cite{mukhoti2020calibrating}. The new layer is trained with early stopping and focal loss which help solve the problem of overfitting and subdued activations, respectively. Early stopping does not allow the layer to overfit the negative log-likelihood loss and hence results in better calibration. Note that any loss of performance due to early stopping is not important as it is not supposed to be used for inference. Focal loss, which is a dynamically weighted variant of the negative log-likelihood loss, helps the network to focus on statically-rare tail-class examples, and hence mitigate the problem of subdued activations. Additionally, focal loss also acts as an entropy-maximizing regularisation which has an added benefit further preventing the model to become overconfident on head-classes \cite{mukhoti2020calibrating}.

Formally, we denote a truncated copy of the model $f_\theta(x)$ with the removed softmax-layer as $\tilde{f}_\theta(x)$, and the newly-initialized recalibration layer, with learnable parameters $\phi$, as $g_\phi$. For an instance $x$ in dataset $D_L$ or $D_U$, the recalibrated activations are obtained as $z = g_\phi(\tilde{f}_\theta(x))$.
The parameters $\theta$ are frozen to prevent any update to the model. The parameters $\phi$ are learned by backpropagation using the focal loss \cite{lin2017focal} which can be written as $FL(p_t) = -\alpha_t(1-p_t)^\gamma \log(p_t)$; where $p_t$ is the probability corresponding to the ground truth class $t$, $\alpha_t \in [0,1]$ is the weighting factor for the $t^{th}$ class, usually set equal to the inverse class frequencies, and $\gamma \ge 0$ is the tunable focussing parameter. Refer to \cite{lin2017focal} for more details. For each example $x$, the probabilities $p_t$ are obtained by taking a softmax over the logits $z$.

\subsubsection{The Minority-class Mining Autoencoder:}
Minority-class mining autoencoder (MCM-AU) aims to solve the problem of information loss due to the aggregation of class-probability distribution in typical uncertainty sampling (\S5.1). For instance, the entropy baseline aggregates information-rich probability distribution into a scalar metric ($\sum_{k}^C p^i_k \log p^i_k$). Hence, it fails to capture vital information about the pattern, relationships, and interactions among individual probability values. Further, these mathematical formulae are fixed (without any parameters), which, unlike a learning process, do not allow it to adapt to identifying and discriminatory patterns in data. MCM-AU, on the other hand, is a parametric function that learns the characteristic patterns and relationships among individual class-activation values. Specifically, the autoencoder, trained using a skewed training dataset, is expected to overfit the activation pattern of majorly occurring head-class examples. During inference, the activation pattern of a rarely occurring tail-class example acts as an anomaly and is flagged as a minority-class example. The key idea is that a tail-class activation pattern is not expected to be reconstructed accurately by an autoencoder trained majorly on head-class data.

Formally, we denote $A_E:Z\rightarrow W$ as encoder and $A_D:W\rightarrow Z$ as decoder. Given a logit vector $z \in Z$, the encoder maps it to a latent representation $w \in W$, which acts as a bottleneck. The decoder maps the latent representation $w$ to output a potential reconstruction $\hat{z} = A_D(A_E(z; \psi_E); \psi_D)$
where $\psi_E$ and $\psi_D$ denote the learnable parameters, trained by optimising the mean-squared error loss, which can be written as MSE$ = ||z-\hat{z}||^2_2$.

\subsubsection{The Decision Module:}
The main job of the decision module is to rank examples based on their likelihood of belonging to one of the minority classes. As explained earlier, minority-class examples are less likely to be reconstructed well using an autoencoder trained on majorly present head-class examples. For each example $x$, the decision module computes the reconstruction error (MSE) between the softmax of the calibrated logits $z$ and their reconstruction $\hat{z}$. The softmax helps to accentuate the difference and works better in practice. Finally, it sorts all the examples based on their score as: 
{\small
\begin{align}
S_x &= ||\sigma(z)-\sigma(\hat{z})||^2_2, \label{eq:mse}\\ 
\text{Rank List} &:= invsort_{\forall x \in D_U} (S_x). \label{eq:rank}
\end{align}
}
However, in the case of object detection, each example x can have multiple candidate detections. Hence we have multiple sets of logits $z_j$, reconstructed logits $\hat{z_j}$, and an additional score $s_j$, quantifying the confidence of a particular detection $d_j$. We use aggregation and filtering to obtain a single example-level score for each example $x$. Specifically, we obtain a set of most-confident (topK) detections. Then, for each topK detection, we compute the reconstruction error (Eq. \ref{eq:mse}) and aggregate them by averaging. 

\section{Evaluation}
\subsection{Datasets and setup} \label{sec:setup}
We conduct experiments on three different datasets, covering two computer vision tasks. The details and relevant summary statistics are presented in Table \ref{table:datasets}. The COCO-11 and SIFAR-10 are derived from benchmark object detetcion and classification datasets, namely MS-COCO \cite{lin2014microsoft} and CIFAR-10 \cite{krizhevsky2009learning}, respectively. CakeDet is a real-application dataset, on cake detection, sourced from one of our high-end clientele, who is struggling to gather labeled images for a rare item in their inventory, i.e. \textit{chocolate}. For each dataset, the tail classes are identified as poorly-performing classes with skewness-ratio less than 0.30, where  \textit{skewness-ratio} ($s$) is defined as the ratio between the number of samples in the particular class divided by the mean number of samples across all classes. While COCO-11 and CakeDet have a similar skewness, SIFAR-10, on the other hand, is designed to be significantly more skewed (see Table \ref{table:datasets}). 

The sub-datasets Train, Pool, and Test (Table \ref{table:datasets}) correspond to different sub-samples of a particular data, where the set Train is used to train various models (the initial skewed target model, re-calibration layer, and autoencoder), the Pool is used to mine minority-class examples and evaluate the mining algorithm's performance, and lastly, the Test is used to test the final performance of the final finetuned model. Unlike CakeDet,  CIFAR-10 and MSCOCO datasets are not naturally skewed. To simulate the skewed experimental setup, we obtain the above-mentioned three separate stratified subsets. We artificially induce a long-tail pattern of skewness in the Train set (corresponding datasets and tail-classes are shown in Table \ref{table:datasets}). We keep the Pool and Test to be uniform distributions. This allows better analysis, interpretation, and comparison of results among individual classes. However, in the case of real-application CakeDet, all subsets Train, Pool, and Test are similarly skewed.

\begin{table*}[t]
\centering
\caption{{Statistics of the various datasets used in our experiments. Tail-class skewness-ratio {(\S\ref{sec:setup})} is based on the train sub-dataset. COCO-11 (11/80 categories) and SIFAR-10 are sub-samples of MS-COCO~\cite{lin2014microsoft} and CIFAR-10~\cite{krizhevsky2009learning} datasets, respectively.}}
\begin{center}
\begin{tabular}{|l||l|l|l|l|l|l|}
\hline
Dataset & \#Class & Tail class names (skewness-ratio) & \#Train & \#Pool & \#Test \\ \hline \hline
COCO-11  & 11 & car (0.05), chair (0.14), bottle (0.23) & 4594  & 7445  & 3723  \\ 
SIFAR-10  & 10 & airplane (0.003), automobile (0.04)     & 14398 & 24000 & 12000 \\ 
CakeDet & 31 &chocolate (0.04)                  & 4091  & 11877 & 650  \\ \hline
\end{tabular}
\end{center}
\label{table:datasets}
\end{table*}

\noindent
\textbf{Evaluation:}
In order to evaluate, each mining algorithm draws candidate examples from the pool set, which are expected to belong to one of the minority classes (Table \ref{table:datasets}). To quantify the mining performance, we use Precision, Precision-Recall (PR) curve, Area Under Precision-Recall curve (AUC-PR), and average F-score. Precision $P$, computed as $\frac{TP}{TP+FP}$, quantifies the percentage of minority-class examples in a queried sample of a certain size. Another quantity of interest is recall, computed as $\frac{TP}{TP+FN}$, and since there is a trade-off, we can visualize precision at various recalls in the PR curve. Finally, we summarise the PR characteristics in a single metric using AUC-PR, which is the area under the PR curve, and average F-score, which is the average of F-score, computed as $\frac{2P\cdot R}{P+R}$ for all points in the PR curve. TP represents true positive, FP is false positive and FN is false negative.

After mining minority-class examples, we finetune the model with the new training dataset and measure improvement in performance on the minority classes. In the case of SIFAR-10, we use classification accuracy (CAcc), and in the case of COCO-11 and CakeDet we use MS-COCO object-detection metrics \cite{lin2014microsoft} (Average Precision(AP) and Average Recall(AR)). Since our algorithm performs significantly better than all the baselines, for better interpretation and conciseness, we also report relative percentage improvement, which is computed as $\frac{(R_{our} - R_{baseline})}{R_{baseline}}*100$, where $R_{our}$ and $R_{baseline}$ represent performance of our approach and a particular baseline, respectively. This quantifies the percentage improvement of our approach with respect to other baselines.

\noindent
\textbf{Baselines:}
We compare our approach with multiple state-of-the-art baselines. 1) Random (samples randomly), 2) MaxScore ($\phi_{MS}^i = 1 - \max_{k} p^i_k$) \cite{culotta2005reducing}, 3) Entropy ($\phi_{ENT}^i = -\sum_{k}^C p^i_k \log p^i_k$) \cite{shannon2001mathematical}, 4) Weighted Shannon Entropy ($\phi_{WENT}^i = -\sum_{k}^C \frac{p^i_k}{b_k.C} \log \frac{p^i_k}{b_k.C}$) \cite{kirshners2017entropy}, and 5) HAL (exploration based) \cite{kazerouni2020active}.
$p_k^i$ is softmax probability for example $i$ and $b_k$ is the proportion of samples corresponding to the $k^{th}$ class among $C$ total classes. Additionally, we tried USBC \cite{chen2011active}, which is a query-by-committee based approach geared towards skewed datasets and a recent approach designed for multi-class imbalance problems \cite{bhattacharya2019generic}. However, the former performed significantly worse than random, possibly due to large skew across multiple classes and the latter did not scale to the size of our datasets due to the huge memory overhead of solving Eq 7 in \cite{bhattacharya2019generic}. Note that all of these baselines are designed primarily for classification and some are not easily extendable to other tasks. Hence, in the case of object detection datasets, we did not try some of these baselines like \cite{chen2011active} and \cite{bhattacharya2019generic}.

\noindent
\textbf{Implementation details:}
We implement our models and design our experiments in PyTorch. For classification, we used a vanilla 4-layered Convolutional Neural Network (CNN), followed by a Fully-Connected (FC) layer. For object detection, we used a RetinaNet detector with a ResNet18/50 backbone. These networks are trained for 20 epochs with batch sizes of 64 and 6, respectively. Similar to the respective final layer, the re-calibration layer is an FC layer in the case of classification and a CNN layer for object detection. Auto-encoder is composed of a two-layered (with hidden dimensions 10 and 9) fully connected encoder and a similar decoder with ReLU activations. It is trained for 10-40 epochs using MSE loss and a learning rate of $1e^{-3}$. During finetuning, the models are trained using the training dataset augmented with the mined minority-class examples for another 20 epochs. All results are averaged over multiple runs to rule out the effect of randomness. The code of our work is accessible at https://tinyurl.com/yckvyyre.

\begin{figure*}[t]
\centering
\hspace{-3.5mm}
\includegraphics[width=.34\textwidth]{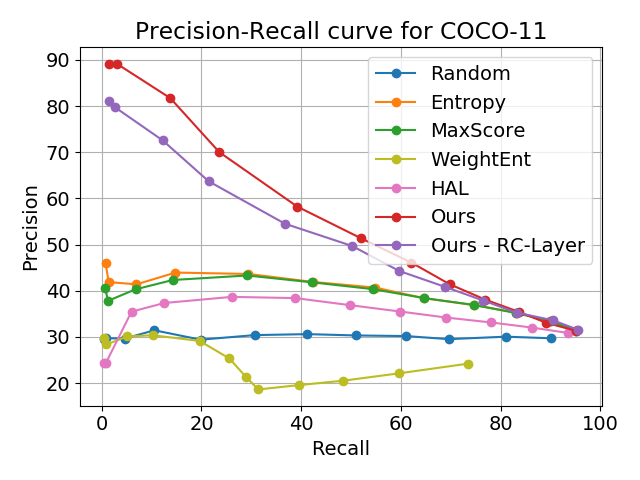}\hspace{-1.3mm}
\includegraphics[width=.34\textwidth]{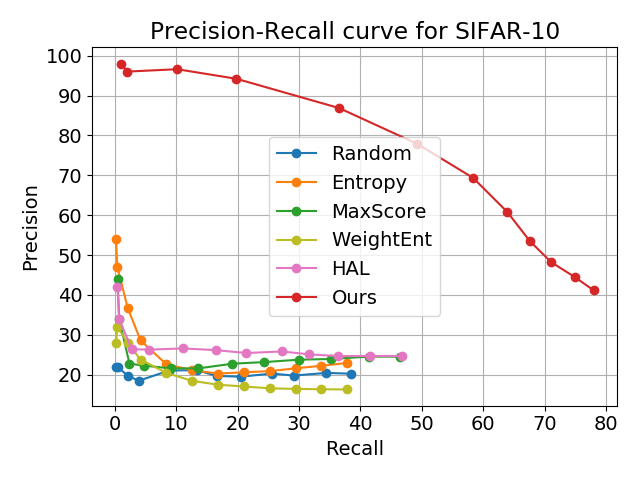}\hspace{-1.3mm}
\includegraphics[width=.34\textwidth]{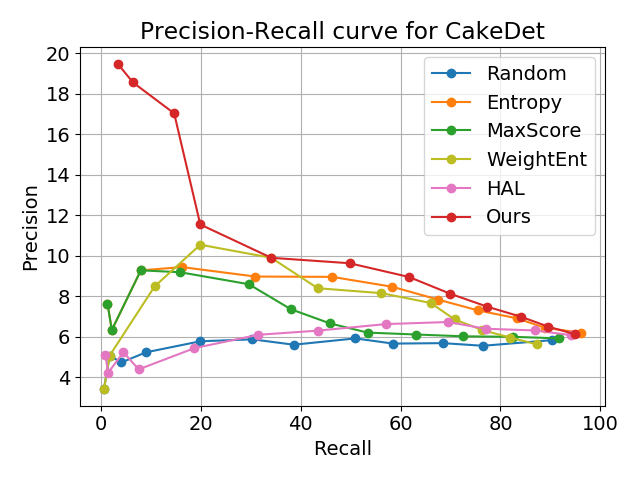}
\caption{Precision-recall characteristics of minority-class mining for different datasets. Our approach outperforms all baselines. Curves towards the top-right are better.}
\label{fig:pr-curves}
\end{figure*}

\subsection{Experiments}
\subsubsection{Minority-class mining performance:} In this experiment, we analyze the minority-class mining performance of our approach in comparison to start-of-the-art baselines.
Specifically, we generate a rank-list (Eq. \ref{eq:rank}) of being a minority-class example (column 3 of Table \ref{table:datasets}) for the entire pool set.  Then, we compute the precision of minority-class mining for sets of various sizes, sampled from the top of the rank list. The corresponding precision-recall curves are provided in Fig \ref{fig:pr-curves}. In terms of relative percentage improvement, for COCO-11 (left), our approach (red) beats all baselines by \textbf{16-44\%} on average F-Score and \textbf{40-200\%} on AUC-PR, depending on the baseline. These figures are \textbf{133-233\%} and \textbf{409-692\%} for SIFAR-10 (middle), and \textbf{17-64\%} and \textbf{21-87\%} for CakeDet (right). 
Significantly improved performance of our approach demonstrates the value of our autoencoder-based approach.
Notice that, the percentage improvement is significantly higher for the highly-skewed SIFAR-10. We suspect this is due to the fact that usual uncertainty sampling baselines degrade under high skewness and perform as worse as random, as concluded in the previous research \cite{attenberg2010label,attenberg2011inactive}. Our approach, on the other hand, does not suffer from any such degradation. Slightly worse performance of all baselines on COCO-11, we believe, is due to noise introduced by filtering and aggregation (\S 3) for object detection problems. Also for CakeDet, all methods perform relatively poorly since here, unlike the other two datasets, the Pool set is also class-imbalanced.

\subsubsection{Robustness of hypothesis}
\begin{figure*}[b!]
\vspace{-4mm}
\centering
\subfigure[Trend of F-Score vs No of minority classes]{
    \centering
    \includegraphics[width=0.42\textwidth]{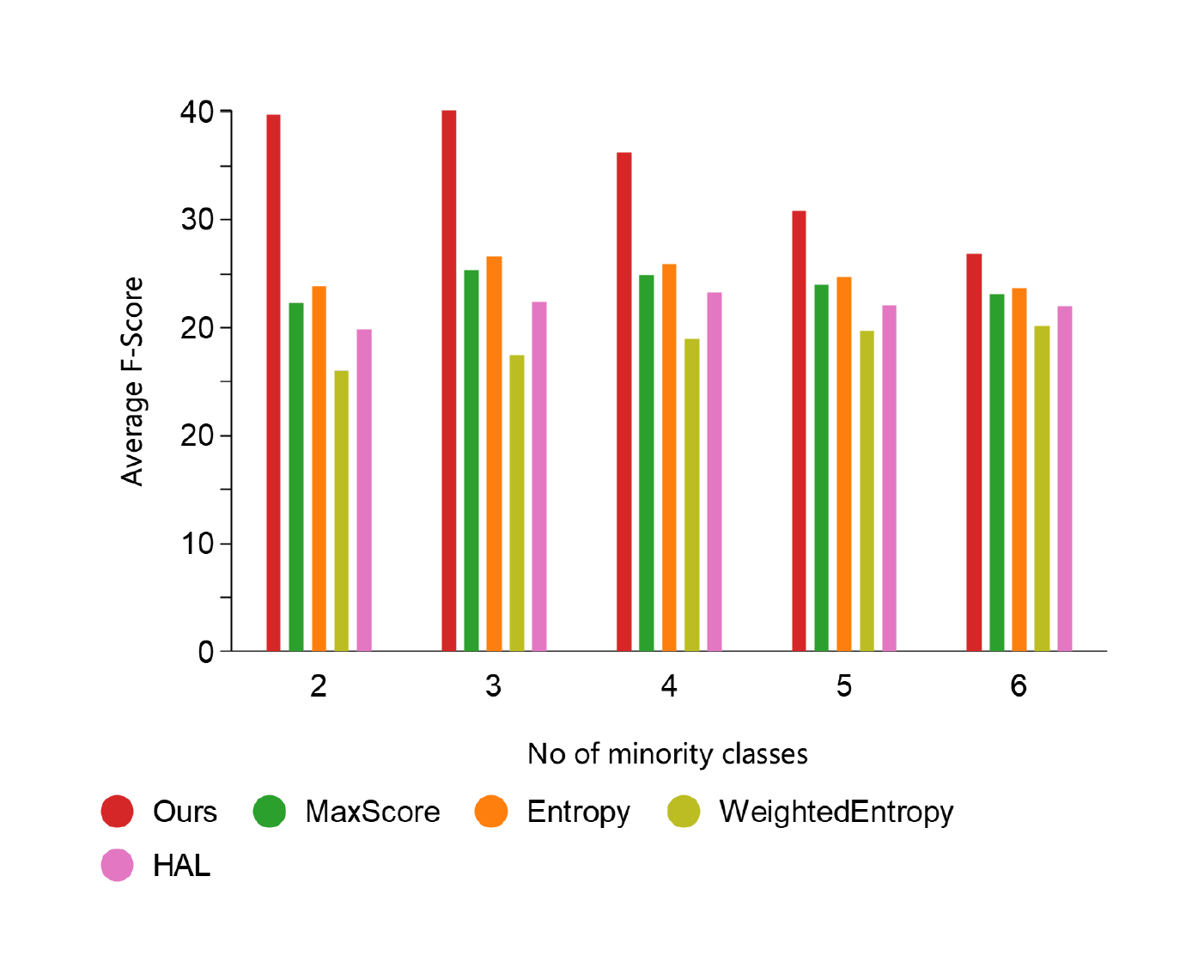}
}\quad
\subfigure[Trend of F-Score vs Different minority classes]{
    \centering
    \includegraphics[width=0.42\textwidth]{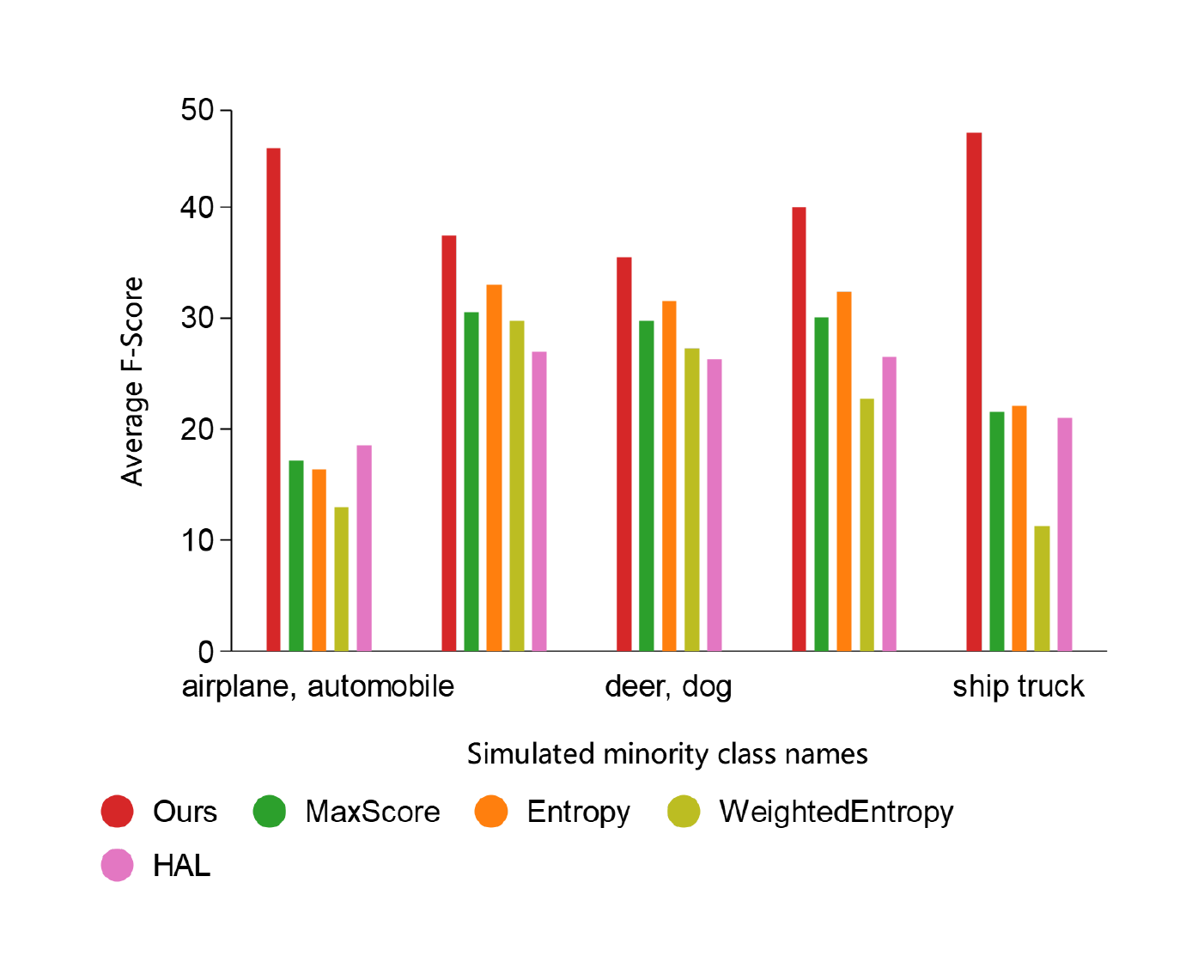}
    }
\caption{Shows average F-Score of different baselines on SIFAR-10 as we vary (instead of what is shown in Table 1) \textit{(a)} number of minority classes, and \textit{(b)} choosing different classes. Our approach consistently outperforms all baselines across all settings.}
\vspace{-2mm}
\label{fig:fscore_analysis}
\end{figure*}

: 
In this experiment, we simulate different settings by varying the number of minority classes and also changing the actual minority classes for SIFAR-10 dataset. As shown in Fig \ref{fig:fscore_analysis} (left), our approach continues to outperform all baselines, which eventually becomes less significant as the number of minority classes increases. This is expected as the odds of an example falling into one of the minority classes eventually takes over as the proportion of minority classes rises. Similarly, as shown in Fig \ref{fig:fscore_analysis} (right), the trend generalizes for different sets of minority classes than the one used in the experiment above.

\subsubsection{Fine-tuned Performance:}
In this experiment, we demonstrate the effectiveness of our approach in terms of the performance of the fine-tuned model on minority classes. Here we only compare with MaxScore and Entropy since they show competitive performance in comparison to all other baselines, an observation consistent with previous research \cite{attenberg2011inactive,ramirez2017active}.  Table \ref{tab:finetune} shows the relative-percentage improvement of our method over the baselines for various sample sizes, datasets, and tail classes.  Our approach beats all the baselines by a significant margin, which is up to \textbf{18-21\%} for COCO-11, \textbf{88-117\%} for SIFAR-10, and \textbf{3\%} for CakeDet. For CakeDet, although, our approach mines 10-20\% more tail-class examples 
, we suspect the modest 3\% rise is due to low variance of images (top-view with similar cakes and backgrounds), resulting in diminishing returns after a point.

\begin{table}[t]
    \caption{Relative \% increase in the finetune performance of our approach in comparison to the baselines for \textit{(left)} various sample sizes and \textit{(right)} for tail classes, arranged by decreasing skewness. Results are averaged over multiple runs and sample sizes.}
    
    \begin{minipage}{.5\linewidth}
      \centering
        \scriptsize
        \begin{tabular}{|l||l|l|l|l|}
        \hline
        Dataset & \#Sample & Random  & Entropy & MaxSc. \\ \hline\hline
        \multirow{3}{*}{COCO-11}    & \textbf{300}      & \textbf{46.94\%}   & \textbf{18.62\%}   & \textbf{21.80\%}    \\
                                    & 744      & 29.64\%   & 9.63\%    & 8.37\%     \\
                                    & 1500     & 23.79\%   & 9.20\%    & 9.36\%     \\\hline
        \multirow{3}{*}{SIFAR-10}   & \textbf{212}      & \textbf{82.56\%} & \textbf{88.97\%} & \textbf{117.3\%} \\
                                    & 2000     & 51.21\% & 47.68\% & 46.19\%  \\
                                    & 5000     & 9.56\%  & 31.91\% & 24.74\%  \\\hline
        \multirow{2}{*}{CakeDet}    & \textbf{200}      & \textbf{7.38\%}    & \textbf{3.38\%}    & \textbf{3.31\%}     \\
                                    & 800      & 4.36\%    & 3.38\%    & 3.24\%    \\\hline
        \end{tabular}
    \end{minipage}%
    \begin{minipage}{.5\linewidth}\vspace{-0.85cm}
        \scriptsize
        \begin{tabular}{|l||l|l|l|l|}
        \hline
        Dataset                   & ClassName                 & Random & Entropy & MaxSc. \\ \hline \hline
        \multirow{2}{*}{SIFAR-10} & airplane       & 305.2\% & 66.34\%   & 155.5\%   \\ 
                                  & automobile      & \underline{30.57\%}  & \underline{74.67\%}   & 69.86\%    \\ \hline
        \multirow{3}{*}{COCO-11}  & car             & 34.31\%  & \underline{3.39\%}    & \underline{5.34\%}     \\ 
                                  & chair           & 26.73\%  & 37.34\%   & 36.07\%    \\ 
                                  & bottle          & 7.93\%   & 10.47\%   & 10.66\%    \\ \hline
        \end{tabular}
    \end{minipage} 
\label{tab:finetune}
\end{table}

\begin{figure*}[b!]
\centering
\subfigure[Influence of sample size]{
    \centering
    \includegraphics[width=0.42\textwidth]{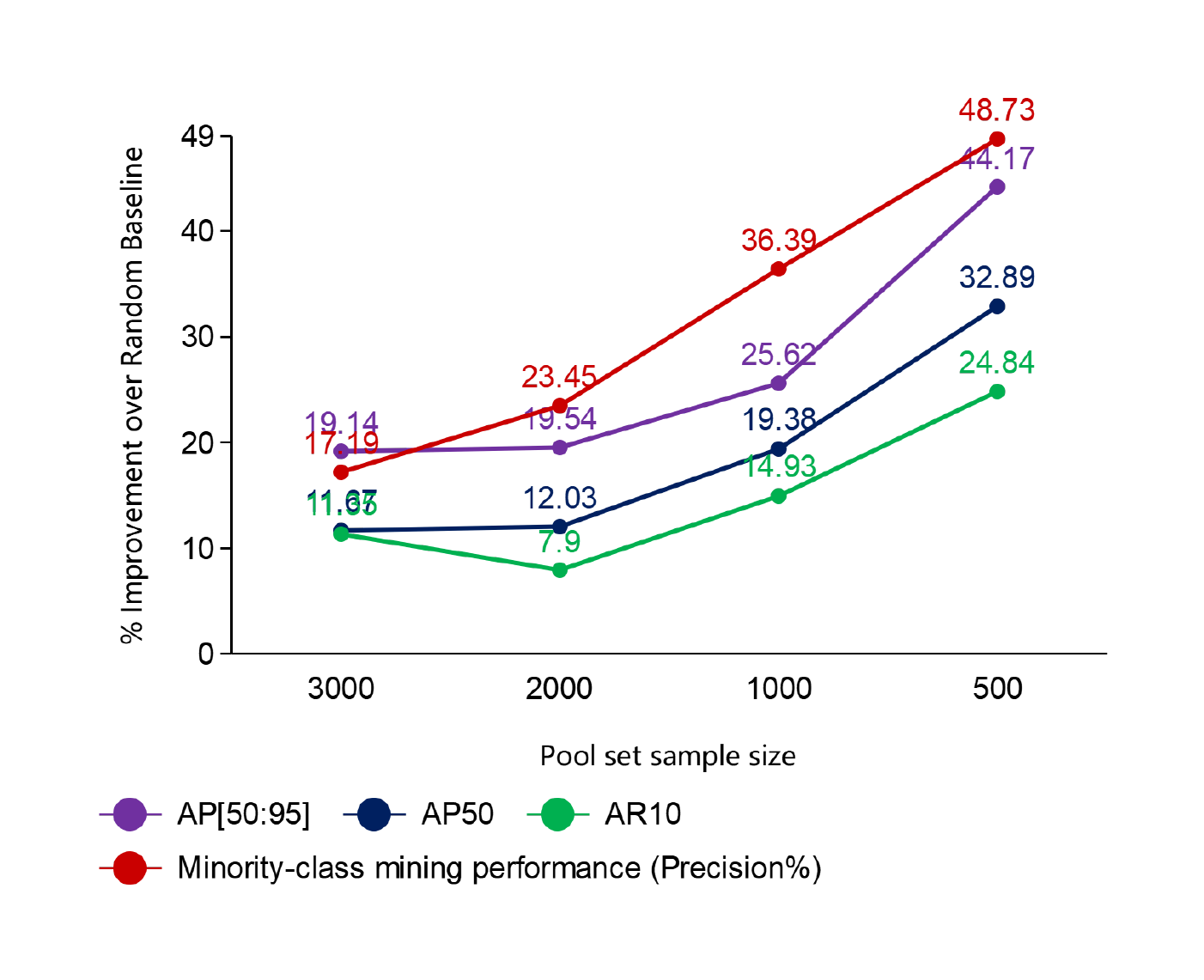}
}\quad
\subfigure[Trend of skewness]{
    \centering
    \includegraphics[width=0.42\textwidth]{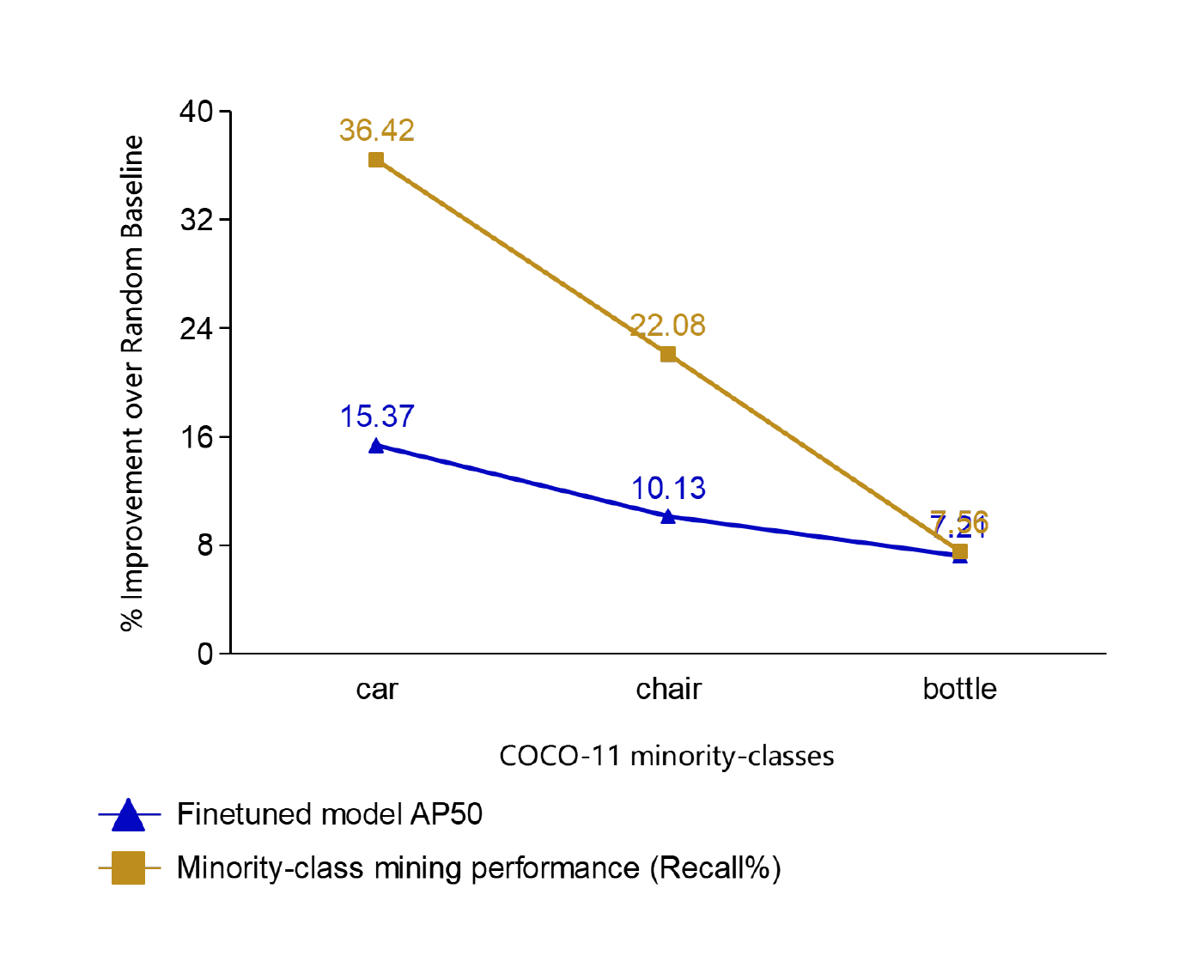}
    }
\vspace{1mm}
\caption{\% improvement for \textit{(a)} various sample sizes and \textit{(b)} tail classes (skewness-ordered). Shows that performance is better for smaller sample and higher skewness.}
\vspace{-3mm}
\label{fig:listanalysis_class}
\end{figure*}

\subsubsection{Further Analysis}
Here, we demonstrate the two desirable properties of the rank list generated by our approach. \textit{First}, the proportion of minority-class examples increases as we move up in the rank list. Fig \ref{fig:listanalysis_class} (a), shows percentage improvement (mining and finetuning) of our approach over the random baseline with different sample sizes for COCO-11. As shown, the mining performance (red) rises rapidly with a lower sample size. Similarly, this trend also generalizes to all baselines and datasets and can also be observed in Table \ref{tab:finetune} (left). \textit{Second}, our approach ranks a greater number of examples belonging to the most skewed classes (hardest examples), up in the rank list. In other words, the most-needed examples (most skewed classes) are prioritized over less-needed examples (less skewed classes). For COCO-11, we show in Fig \ref{fig:listanalysis_class} (b), \textit{car} which is the most skewed class, shows the highest percentage of improvement (mining and finetuning) over the random baseline, which gradually reduces for \textit{chair} and \textit{bottle} as the skewness becomes less severe. 
The same trend can also be observed in fine-tuned model's performance (with a few exceptions, marked with underline) for individual minority classes in Table \ref{tab:finetune} (right), where the classes are arranged with decreasing skewness (from \textit{airplane} to \textit{bottle}).
Observing these trends, we conclude that a higher proportion and high quality (corresponding to most skewed classes) of minority-class examples are ranked higher in our approach in comparison to other baselines. This is a very desirable property for finetuning the model with baby steps, like in active learning, instead of one large step.

\subsubsection{Ablation Study}
Finally, we conduct an ablation study to clearly understand the value addition of individual components of our framework (i.e. RC-layer and the one-class autoencoder). In the PR-curve for COCO-11 (Fig \ref{fig:pr-curves} left), \textit{Ours} (red) represents the performance of our approach and \textit{Ours - RC-Layer} (purple) represents the performance of our approach without the RC-Layer. The importance of recalibrating activations is established by the significant drop in performance after ablating the RC-Layer. Further, note that the only difference between Ours - RC-Layer (purple) and uncertainty baselines (green and orange) is the use of autoencoder instead of uncertainty metrics. Substantial improvement (purple vs. green and orange) demonstrates the importance of using an autoencoder instead of a fixed formula.

\section{Conclusion}
This study presents a novel, simple, and effective framework to mine minority-class examples using uncertainty estimates. We identified key challenges due to which earlier works struggled to mine minority-class examples. We addressed these challenges using one-class modeling of re-calibrated activations. Consequently, we demonstrate the effectiveness of our method by a significant improvement in minority-class mining and fine-tuned model performance. Additionally, we demonstrated two desirable properties of our approach to mine better samples, both in terms of quality and quantity, up in the rank list.

\bibliographystyle{splncs04}
\bibliography{main}
\end{document}